\documentclass[10pt,twocolumn,letterpaper]{article}

\usepackage{cvpr}

\usepackage[utf8]{inputenc} 
\usepackage[T1]{fontenc}    
\usepackage[pagebackref=true,breaklinks=true,letterpaper=true,colorlinks,bookmarks=false]{hyperref}
\usepackage{url}            
\usepackage{booktabs}       
\usepackage{amsfonts}       
\usepackage{nicefrac}       
\usepackage{microtype}      

\usepackage{times}
\usepackage{epsfig}
\usepackage{graphicx}
\usepackage{amsmath}
\usepackage{amssymb}
\usepackage{subfigure}
\usepackage{algorithm}
\usepackage{algorithmic}
\usepackage{placeins}
\usepackage{array}
\usepackage{multirow}

\cvprfinalcopy 

\def\cvprPaperID{1693} 

\ifcvprfinal\fi

\begin{document}

\title{Learning Convolutional Neural Networks using Hybrid Orthogonal Projection and Estimation}

%

\author{Hengyue Pan\\
York University\\
4700 Keele Street, Toronto, Ontario, CA\\
{\tt\small panhy@cse.yorku.ca}\\
\and
Hui Jiang\\
York University\\
4700 Keele Street, Toronto, Ontario, CA\\
{\tt\small hj@cse.yorku.ca}
}

\maketitle

\begin{abstract}
Convolutional neural networks (CNNs) have yielded the excellent performance in a variety of computer vision tasks, where CNNs typically adopt a similar structure consisting of convolution layers, pooling layers and fully connected layers. In this paper, we propose to apply a novel method, namely Hybrid Orthogonal Projection and Estimation (HOPE), to CNNs in order to introduce orthogonality into the CNN structure. The HOPE model can be viewed as a hybrid model to combine feature extraction using orthogonal linear projection with mixture models. It is an effective model to extract useful information from the original high-dimension feature vectors and meanwhile filter out irrelevant noises. In this work, we present two different ways to apply the HOPE models to CNNs, i.e., {\em HOPE-Input} and {\em HOPE-Pooling}. For {\em HOPE-Input}, a HOPE layer is directly used right after the input to de-correlate high-dimension input feature vectors. Alternatively, in {\em HOPE-Pooling}, a HOPE layer is used to replace the regular pooling layer in CNNs. The experimental results on both CIFAR-10 and CIFAR-100 data sets have shown that the orthogonal contraints imposed by the HOPE layers can significantly improve the performance of CNNs in these image classification tasks (we have achieved top-3 performance when image augmentation has not been applied).
\end{abstract}

\section{Introduction}
\label{Intro}

Convolutional neural networks (CNNs) \cite{le1990handwritten} currently play an important role in the deep learning and computer vision fields. In the past several years, researchers have revealed that CNNs can give the state-of-the-art performance in many computer vision tasks, especially for image classification and recognition tasks \cite{NIPS2012_AlexNet,GoogleLeNet_2014,VGG_2015}. Comparing with the fully connected deep neural networks (DNNs), CNNs are superior in exploring spacial constraints and in turn extracting better local features from input images using the convolution layers and weight sharing, and furthermore may provide better invariance through the pooling mechanism.
All of these make CNNs very suitable for image-related tasks \cite{lecun1995convolutional}. Moreover, large-scale deep CNNs can be effectively  learned end-to-end in a supervised way from a large amount of labelled images.

In the past several years, a tremendous amount of research effort has been devoted to further improve the performance of deep CNNs. In \cite{hinton2012improving,srivastava2014dropout}, the dropout method has been proposed to prevent CNNs from overfitting by randomly dropping a small portion of hidden nodes in the network during the training procedure. Many experiments have confirmed that the dropout technique can significantly improve the network performance, especially when only a small training set is available. Besides, a similar idea, called dropconnect \cite{wan2013dropconnect}, has been proposed to drop connections between layers instead of hidden nodes during the training stage. Another interesting research field is to design good nonlinear activation functions for neural networks beyond the popular rectified linear function (ReLU), such as maxout \cite{goodfellow2013maxout} and PReLU \cite{he2015delving}, which are also demonstrated to yield improvement in terms of classification performance. On the other hand, another important path to improve model  performance is to search for some new CNN structures. For example, in \cite{lin2013network}, Network in Network (NIN) has been proposed, in which one micro neural network is used to replace the regular linear convolutional filter. Recurrent Convolutional Neural Network (R-CNN) \cite{liang2015recurrent} is another new CNN structure, which introduce recurrent connections into the convolution layers. In \cite{rippel2015spectral}, the spectral pooling method is proposed, which applies discrete Fourier transform into the pooling layers to preserve more useful information after the dimensionality reduction.


More recently, a novel model, called Hybrid Orthogonal Projection and Estimation (HOPE) \cite{zhang2015hybrid}, has been proposed to learn neural networks in either supervised or unsupervised ways. This model introduces a linear orthogonal projection to reduce the dimensionality of the raw high-dimension data and then uses a finite mixture distribution to model the extracted features. By splitting the feature extraction and data modeling into two separate stages, it may derive a good feature extraction model that can generate better low-dimension features for the further learning process. More importantly, based on the analysis in \cite{zhang2015hybrid}, the HOPE model has a tight relationship with neural networks since each hidden layer of DNNs can also be viewed as a HOPE model being composed of the feature extraction stage and data modeling stage. Therefore, the maximum likelihood based unsupervised learning as well as the minimum cross-entropy error based supervised learning algorithms can be used to learn neural networks under the HOPE framework for deep learning. In this case, the standard back-propagation method may be used to optimize the objective function to learn the models except that the orthogonal constraints are imposed for all projection layers during the training procedure.

However, \cite{zhang2015hybrid} has not taken CNNs into account but merely investigated the HOPE models for the fully connected neural networks and demonstrated good performance in the small MNIST data set. In this paper,  we extend the HOPE model to the popular CNNs by considering the special model structures of both convolution and pooling layers, and further consider how to introduce the orthogonal constraints into the CNN model structure and learn CNNs under the HOPE framework. The most straightforward idea is to use a HOPE layer as the first hidden layer in CNNs to de-correlate the high-dimension input CNN features and remove the irrelevant noises as a result, which we call HOPE-Input layer. This idea is similar as the original formulation in \cite{zhang2015hybrid} except the HOPE model is applied to each convolutional filter.
Moreover, 
the pooling layers, using either average pooling or max pooling, are a critical step in CNNs \cite{jarrett2009best} since they can reduce the resolution of the lower-level feature maps and then make the models more tolerable to the slight distortion or translation in the orignal images \cite{le1990handwritten}. In \cite{boureau2010theoretical}, a theoretical analysis of average pooling and max pooling is made to reflect how pooling can affect the network performance. However, in most cases, the pooling layers are still used based on empirical information. In \cite{springenberg2014striving}, it proposes a new CNN structure using larger stride convolution layers to replace the pooling layers, and the authors argue that the larger stride convolution layers can perform equally well as the pooling layers and also achieve similar performance in the experiments. Hinted by this idea, we propose another method to apply the HOPE models to CNNs, namely using the HOPE models to replace the regular pooling layers, called a HOPE-Pooling layer. In this way, the orthogonality is further introduced to the models in this stage.
Our experimental results on both CIFAR-10 and CIFAR-100 data sets have shown that the both HOPE-Input layers and HOPE-Pooling layers result in significant performance improvement over the regular CNN baseline models \footnote{The code of our HOPE CNN can be downloaded via: https://github.com/mowangphy/HOPE-CNN}.

The structure of the rest of this paper is listed below:  section~\ref{HOPE} will briefly review the  HOPE model and its usage in DNNs. In section~\ref{Method}, we present both HOPE-Input layers and HOPE-Pooling layers.  In section~\ref{Experiments}, we report experimental results on two popular data sets, namely CIFAR-10 and CIFAR-100, and compare with other CNN models.  Finally, in section~\ref{Conclusion}, we conclude the paper with the findings and conclusions.

\section{Hybrid Orthogonal Projection and Estimation (HOPE) Framework}
\label{HOPE}

In the original Hybrid Orthogonal Projection and Estimation (HOPE) formulation \cite{zhang2015hybrid}, it is assumed that
any high-dimension feature vector can be modelling by a hybrid model consisting of feature extraction using a linear orthogonal projection and statistic modeling using a finite mixture model.
Assume each high-dimension feature vector ${\bf x}$ is of dimension $D$, the linear orthogonal projection will map ${\bf x}$ to an $M$-dimension feature space ($M < D$), and the projected vector may attain the most useful information of ${\bf x}$. Specifically, we can define a $D \times D$ orthogonal matrix $[ {\bf U} ; \;\; V ]$ which satisfies:
\begin{equation}
\label{eq-signal-noise}
[{\bf z}; \;\; {\bf n}] = [ {\bf U} ; \;\; V ] \; {\bf x}
\end{equation}
where ${\bf z}$ is an $M$-dimension vector, called the signal component, and ${\bf n}$ is the residual noise vector with the dimensionality of $D - M$.

In practice, ${\bf z}$ is heavily de-correlated but it may still locate in a rather high dimension feature space. In the HOPE formulation, it is proposed to model ${\bf z}$ with a finite mixture model:
\begin{equation}
\label{eq-signal-model}
p({\bf z}) = \sum_{k=1}^K \pi_k \cdot f_k({\bf z}|\theta_k)
\end{equation}
where $K$ is the number of mixture components, $\pi_k$ is the mixture weight of the $k$th component ($\sum_{k=1}^K \pi_k = 1$), $f_k()$ denotes a selected distribution from the exponential family, and $\theta_k$ denotes all model parameters of $f_k()$. As discussed in \cite{zhang2015hybrid}, if the von Mises-Fisher (vMF) distribution is chosen for $f_k()$, the resultant HOPE model is equivalent in mathematical formulation to a hidden layer in neural networks using the popular rectified linear units (ReLU).

%

\begin{figure}[ht]
\begin{center}
    \includegraphics[width=0.8\linewidth]{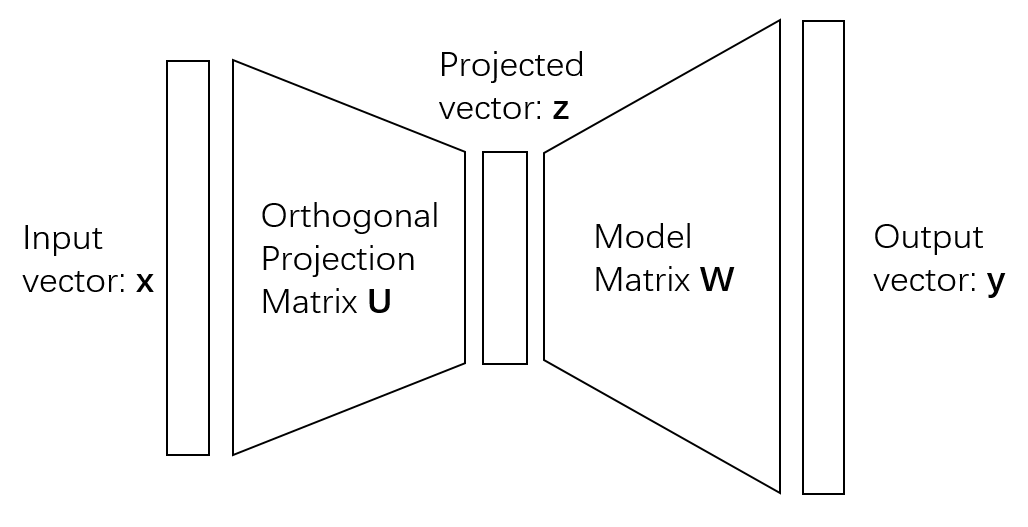}
\end{center}
\caption{The HOPE model is viewed as a hidden layer in DNNs.}
\label{Fig:HOPEDNN}
\end{figure}

The HOPE model combines a linear orthogonal projection and a finite mixture model under a unified generative modeling framework. It can be learned unsupervisely based on the maximum likelihood estimation from unlabelled data as well as discriminatively from labelled data.
In \cite{zhang2015hybrid},  the HOPE model has been applied to the fully connected DNNs and learn the models accordingly in either supervised or unsupervised ways. For one hidden layer with input vector ${\bf x}$ $({\bf x} \in R^D)$ and output vector ${\bf y}$ $({\bf y} \in R^G)$, it is first splited into two layers: i) The first layer is a linear orthogonal projection layer, which is used to project ${\bf x}$ to a feature vector ${\bf z}$ $ ({\bf z} \in R^M, M < D)$ and remove the noise signals by using an orthogonal projection matrix ${\bf U}$:
\begin{equation}
\label{eq-DNN-projection}
{\bf z} = {\bf U}{\bf x}.
\end{equation}
ii) The second layer is a non-linear model layer, which convert ${\bf z}$ to the output vector ${\bf y}$ following the selected model $f_k()$ and a nonlinear log-likelihood pruning operation. An example of a HOPE layer in DNNs is shown in Figure \ref{Fig:HOPEDNN}.

As in \cite{zhang2015hybrid}, all HOPE model parameters, including the projection matrix $U$ and the model matrix $W$, can be learned, using the error back-propagation algorithm with stochastic gradient descent, to optimize an objective function subject to an orthogonal constraint,  ${\bf U} {\bf U}^T = {\bf I}$, for each projection layer. As in \cite{zhang2015hybrid}, for computational simplicity, the constraint is cast as the following penalty term to gradually de-correlate the matrix ${\bf U}$ during the learning process:

\begin{equation}
\label{eq-Upenalty}
P({\bf U}) = \sum_{i=1}^{M} \sum_{j=i+1}^{M} \frac{|{\bf u}_i \cdot {\bf u}_j|}{|{\bf u}_i| \cdot |{\bf u}_j|}.
\end{equation}

In \cite{zhang2015hybrid}, both unsupervised learning and supervised learning are studied for DNNs under the HOPE framework. The above orthogonal constraint is found to be equally important in both scenarios. In this paper, we will study how to supervised learn CNNs under the HOPE formulation and more specifically investigate how to introduce the orthogonality into the CNN model structure.

\section{Our Proposed Method}
\label{Method}

In \cite{zhang2015hybrid}, the authors have applied the HOPE model to the fully connected DNNs and have achieved good performance in experiments on small data sets like MNIST. However, more widely used neural models in computer vision, i.e. convolutional neural networks (DCNNs), have not been considered. Unlike DNNs, CNNs adopt some unique model structures and have achieved huge successes in many large-scale image classification tasks. Therefore, it is interesting to consider how to combine the HOPE model with CNNs to further improve image classification performance.

\subsection{Applying the HOPE model to CNNs}

\begin{figure*}[ht]
\begin{center}
    \includegraphics[width=0.7\linewidth]{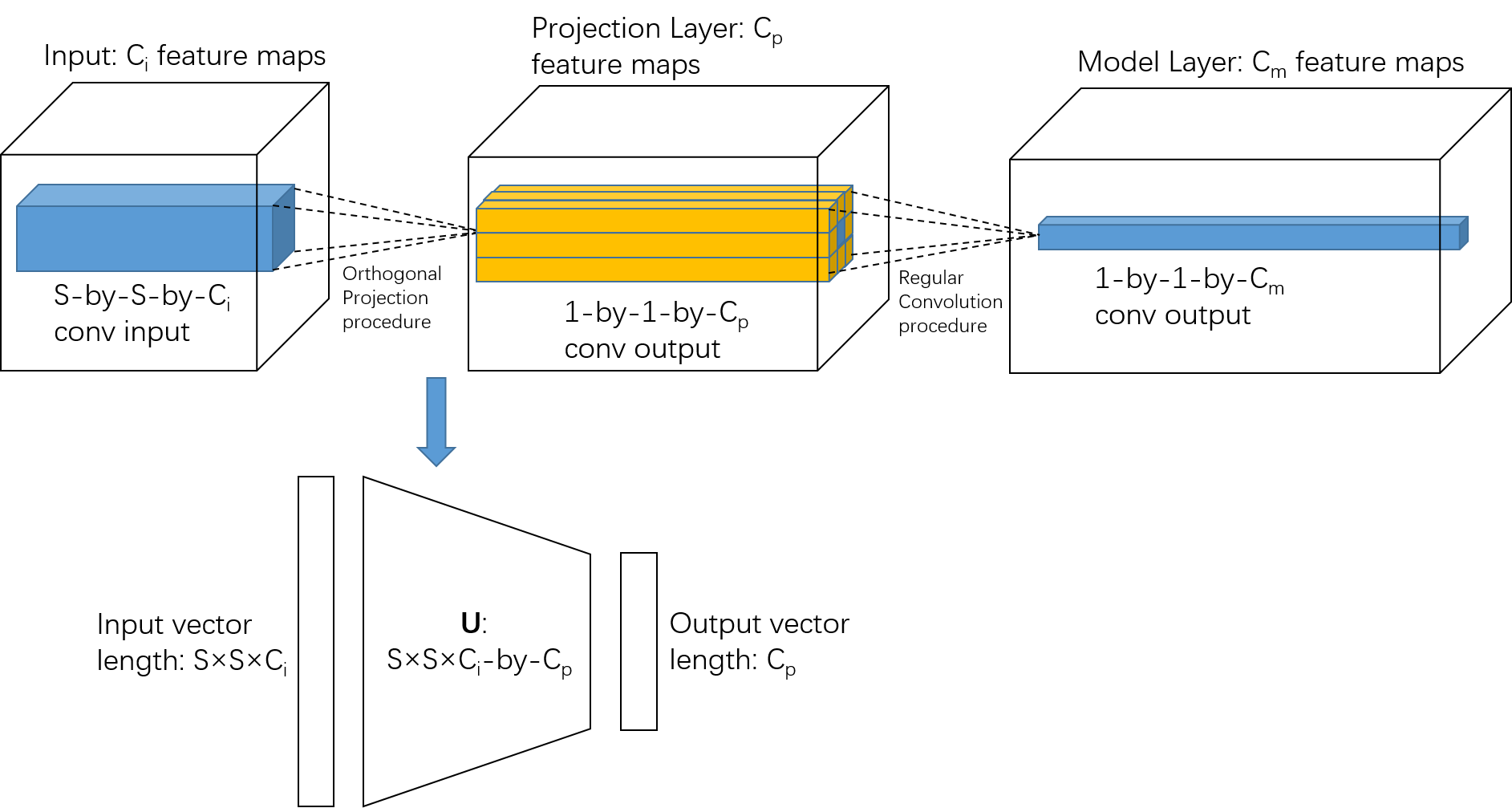}
\end{center}
\caption{A convolution layer in CNNs may be viewed as a HOPE model.}
\label{Fig:Conv2DNN}
\end{figure*}

To apply the HOPE model to CNNs, the most straightforward solution is
to split each convolution layer into a concatenation of a projection layer and a model layer and
impose the orthogonal constraints onto the projection layer as in \cite{zhang2015hybrid}.
Assume that we have a regular convolution layer in CNNs, which uses some $S \times S$ linear filters to map from $C_i$ input feature maps to $C_m$ output feature maps. As shown in Figure \ref{Fig:Conv2DNN}, under the HOPE framework, we propose to split this convolution layer into the two separate layers:

\begin{enumerate}
\item[i)]  One linear orthogonal projection layer with the projection matrix ${\bf U}$: it linearly maps a 3-dimension tensor with the size of $S \times S \times C_i$ into a vector $1 \times 1 \times C_p$, $C_p$ denotes the feature maps to be used in this projection layer. As the projection filters convolve with the input layer, it generates a total of $C_p$ feature maps in the projection layer. The projection filter itself is a 4-dimension tensor with the size of $S \times S \times C_i \times C_p$. Based on the definition of the convolution procedure and follow the formulation in \cite{zhang2015hybrid}, we can reshape this 4-dimension tensor as a matrix ${\bf U}$ with the size of $ (S \cdot S \cdot C_i) \times C_p$, as shown in Figure \ref{Fig:Conv2DNN}.

\item[ii)] One model layer with the weight matrix $W$: it has exactly same structure as a regular convolutional layer, which mapping the $C_p$ projected feature maps into $C_m$ output feature maps. Differing from \cite{zhang2015hybrid}, instead of only mapping the projected vector, the proposed model layer here takes all projected vectors within each $S \times S$ region and map all projected features within this region into the final output feature maps. We have found that this modification is critical in CNNs for better performance in image classification.

\end{enumerate}

Figure \ref{Fig:Conv2DNN} shows the whole structure of one HOPE layer in CNNs. Since the projection layer is linear, we may collapse these two layers to derive a normal convolution layer in CNNs. However, as argued in \cite{zhang2015hybrid}, there are many advantages to separate them so as to learn CNNs under the HOPE framework.

Note that $C_p$ is always far less than $S \times S \times C_i$ in the above HOPE formulation, it implies that the orthogonal projection may help to remove irrelevant noises in this step.

In this paper, we only consider the supervised learning of CNNs under the HOPE framework. In this case, the model parameters in the model layer can be learned in the same way as in the convolutional CNNs. However, for the projection layers, we need to impose the orthogonal constraint, ${\bf U} {\bf U}^T = {\bf I}$ during the learning process. Following \cite{zhang2015hybrid}, we cast this constraint as a penalty term in eq. (\ref{eq-Upenalty}).

First of all, we need to derive the gradient of the penalty term $P({\bf U})$ with respect to ${\bf U}$ as follows:
\begin{equation}
\label{eq-grad-U}
\frac{\partial P({\bf U})}{\partial {\bf u}_i} = \sum_{j=1}^{M} (\frac{|{\bf u}_i \cdot {\bf u}_j|}{|{\bf u}_i| \cdot |{\bf u}_j|}) \cdot \bigg((\frac{{\bf u}_j}{{\bf u}_i \cdot {\bf u}_j}) - (\frac{{\bf u}_i}{{\bf u}_i \cdot {\bf u}_j}) \bigg)
\end{equation}
To facilitate the above computation in GPUs, we may equivalently represent the above gradient computation as a matrix form, i.e., essentially a multiplication of the two matrices ${\bf D}$ and ${\bf B}$ as follows:
\begin{equation}
\label{eq-grad-U-mat}
\frac{\partial P({\bf U})}{\partial {\bf U}} = ({\bf D} - {\bf B}){\bf U}
\end{equation}
where ${\bf D}$ is an $M$-by-$M$ matrix of
$ d_{ij} = \frac{\mbox{sign}({\bf u}_i \cdot {\bf u}_j)}{|{\bf u}_i| \cdot |{\bf u}_j|}
$  $(1< i, j <M)$
and ${\bf B}$ is another $M$-by-$M$ diagonal matrix of
$
b_{ii} = \frac{\sum_{j}g_{ij}}{{\bf u}_i \cdot {\bf u}_i}
$
with $ g_{ij} = \frac{|{\bf u}_i \cdot {\bf u}_j|}{|{\bf u}_i| \cdot |{\bf u}_j|}$  $(1< i, j <M)$.

Secondly, we can combine the above $\frac{\partial P({\bf U})}{\partial {\bf U}}$ with the gradient $\Delta {\bf U}$, which is calculated from the objective function:
\begin{equation}
\label{eq-grad-update}
\widetilde{\Delta {\bf U}}= \Delta {\bf U} + \beta \cdot \frac{\partial P({\bf U})}{\partial {\bf U}}
\end{equation}
where $\beta$ is a pre-defined parameter to balance  the orthogonal penalty term. Finally, the projection matrix ${\bf U}$ can be updated as follows:
\begin{equation}
\label{eq-U-update}
{\bf U}^{(n)} = {\bf U}^{(n-1)} - \gamma \cdot \widetilde {\Delta {\bf U}}
\end{equation}
where $\gamma$ is the learning rate for the weight update.
During the learning process, ${\bf U}$ is gradually de-correlated and eventually becomes an orthogonal matrix.

\subsection{HOPE-Input Layers}

The first way to apply the HOPE model to CNNs is to use the above HOPE layer to replace the first convolution layer right after the image pixel input. The HOPE formulation may help to de-correlate the raw image pixel inputs and filter out irrelevant noises in the first place.
%
This is called as one HOPE-Input layer. 
In practice, we may apply more HOPE layers to replace the following convolution layers in CNNs as well.

\subsection{HOPE-Pooling Layers}

In CNNs, the pooling layers \cite{krizhevsky2012imagenet} are traditionally considered as important for good performance. \cite{springenberg2014striving} has shown that the pooling layers result in the reduction of feature dimensionality, which help the CNNs to {\em view} much larger regions of the input feature maps, and generate more stable and invariant high level features. Moreover, \cite{springenberg2014striving} argues to use regular convolution layers with larger stride to replace the pooling layers, and claims to achieve the similar performance as the pooling layers. The particular network structure is  called 'ALL-CNN'.
ALL-CNN models provide a useful idea that we may use the convolution layers with extra parameters in place of the simple pooling layers in CNNs.

As an alternative way to apply the HOPE model to CNNs, we propose to use the HOPE layer in  Figure  \ref{Fig:Conv2DNN} to replace the normal pooling layers in CNNs. Comparing with the regular pooling layers, we believe that the HOPE layer may be advantageous in feature extraction since the linear orthogonal projection may help to de-correlate the input feature maps and generate better features for the upper layers.
Assume we have a regular pooling layer, which takes a 3-dimension tensor $S \times S \times C_p $ from one region in the input and generate a vector with the size of $1 \times 1 \times C_p $ based on the simple pooling operation, such as {\em max}. Normally, we do not have any learnable parameters in the pooling layers. In this paper, we propose to use a linear orthogonal projection layer with a weight matrix of $(S \cdot S \cdot C_p) \times C_p$ in size, to replace the regular pooling layer. The projection matrix will be learned as above to ensure the orthogonality. We call the linear orthogonal projection layer along with the model layer a HOPE-Pooling layer.

In practice, for simplicity, we just use a linear orthogonal projection layer to replace a pooling layer in CNNs, and the convolution layer next to it can be viewed as a model layer. In this way, we can reduce the number of new parameters to be introduced in our formulation. In this case, we still introduce about $(S \cdot S \cdot C_p \cdot C_p)$ more parameters.
To make sure our model is still comparable with the baseline in terms of model size, we only use one HOPE-Pooling layer to replace the first pooling layer in CNNs, which normally has much less feature maps (where $C_p$ is quite small) and keep the other regular pooling layers unchanged. Adding more HOPE layers may result in a much bigger model, which may quickly overfit a small training set.

\begin{table*}[htb]
  \caption{The structure of several CNNs examined in this work}
  \label{table-CNNs}
  \centering
  \begin{tabular}{p{3.5cm}<{\centering}|p{3.5cm}<{\centering}|p{3.5cm}<{\centering}}
    \toprule
    Baseline & HOPE-Input & Single HOPE-Pooling  \\
    \midrule
    \multicolumn{3}{c}{Input: 32-by-32 images in RGB color channel}                   \\
    \midrule
    \multirow{4}*{-} & {\bf $3 \times 3$ filter} & \multirow{4}*{-}            \\
    & {\bf 20 feature maps }         \\
    & {\bf using orthogonal }               \\
    & {\bf projection }                  \\
    \midrule
    \multicolumn{3}{c}{$3 \times 3$ filter, 64 feature maps, batch normalization, ReLU, dropout 0.3}                                \\
    \multicolumn{3}{c}{$3 \times 3$ filter, 64 feature maps, batch normalization, ReLU}                                             \\
    \midrule
    $2 \times 2$ & $2 \times 2$ & {\bf $2 \times 2$ filter}        \\
    max-pooling & max-pooling & {\bf 64 feature maps} \\
    stride = 2  & stride = 2  &  {\bf using orthogonal projection, stride = 2} \\
    \midrule
    \multicolumn{3}{c}{$3 \times 3$ filter, 128 feature maps, batch normalization, ReLU, dropout 0.4}                                \\
    \multicolumn{3}{c}{$3 \times 3$ filter, 128 feature maps, batch normalization, ReLU}                                             \\
    \midrule
    $2 \times 2$  & $2 \times 2$  &  $2 \times 2$                 \\
    max-pooling & max-pooling & max-pooling                    \\
    stride = 2  & stride = 2 & stride = 2                       \\
    \midrule
    \multicolumn{3}{c}{\{$3 \times 3$ filter, 256 feature maps, batch normalization, ReLU, dropout 0.4\} $\times 2$}                                \\
    \multicolumn{3}{c}{$3 \times 3$ filter, 256 feature maps, batch normalization, ReLU}                                             \\
    \midrule
    $2 \times 2$ & $2 \times 2$ &  $2 \times 2$                               \\
    max-pooling & max-pooling & max-pooling                   \\
    stride = 2  & stride = 2 & stride = 2                     \\
    \midrule
    \multicolumn{3}{c}{ \{ $3 \times 3$ filter, 512 feature maps, batch normalization, ReLU, dropout 0.4 \} $\times 2$}                                \\
    \multicolumn{3}{c}{$3 \times 3$ filter, 512 feature maps, batch normalization, ReLU}                                             \\
    \midrule
    $2 \times 2$ & $2 \times 2$ &  $2 \times 2$                                  \\
    max-pooling & max-pooling & max-pooling                              \\
    stride = 2  & stride = 2 & stride = 2                           \\
    \midrule
    \multicolumn{3}{c}{ \{ $3 \times 3$ filter, 512 feature maps, batch normalization, ReLU, dropout 0.4 \} $\times 2$}                                  \\
    \multicolumn{3}{c}{$3 \times 3$ filter, 512 feature maps, batch normalization, ReLU}                                             \\
    \midrule
    $2 \times 2$ & $2 \times 2$ &  $2 \times 2$                              \\
    max-pooling & max-pooling & max-pooling                      \\
    stride = 2  & stride = 2 & stride = 2                          \\
    \midrule
    \multicolumn{3}{c}{Fully connected layer, 512 nodes, batch normalization, ReLU, dropout 0.5}                                \\
    \multicolumn{3}{c}{Fully connected layer, 10 nodes, with softmax} \\
    \bottomrule
  \end{tabular}
\end{table*}

\section{Experiments}
\label{Experiments}

In this paper, we use two widely used image classification data sets, namely CIFAR-10 and CIFAR-100 \cite{krizhevsky2009learning}, to evaluate the performance of our proposed HOPE-Input and HOPE-Pooling methods.

\subsection{Databases}

CIFAR-10 and CIFAR-100 are two popular data sets that are widely used in computer vision.  Both data sets contain 50,000 32-by-32 RGB images for training and 10,000 images for validation. The main difference between these two data sets is that CIFAR-10 only divides all images into 10 coarse classes, but CIFAR-100 divides them into 100 fine classes. In our experiments, we should transform all images into the YUV color channels.

\subsection{Experimental Configurations}

In our experiments, we consider several different CNN structures as specified in Table \ref{table-CNNs} in detail. Firstly, we follow the CNN structure that is defined by Sergey Zagoruyko as our baseline CNNs.\footnote{See https://github.com/szagoruyko/cifar.torch for more information. According to the website, without using data augmentation, the best performance on the CIFAR-10 test set is 8.7\% in error rate. By using RGB color channel instead of YUV, our reproduced baseline performance is 8.30\% in this paper.} Then we evaluate the HOPE-Input CNN and HOPE-Pooling CNN as discussed in Section ~\ref{Method}, and compare them with the baseline model. In Table \ref{table-CNNs},  we have provided the detailed description  of the structure of 3 CNNs (baseline, HOPE-Input and HOPE-Pooling) used in our experiments. Moreover, we also consider another configuration by combining  HOPE-Input and HOPE-Pooling, i.e., using one HOPE-Input layer and one HOPE-Pooling layer at the same time.

To further investigate the performance of the HOPE-Input CNN,  we also consider a model configuration
called as LIN-Input CNN, which uses the same model structure as the HOPE-Input CNN except that the orthogonal constraint in eq. (\ref{eq-Upenalty}) is NOT applied in training. Similarly, for the HOPE-Pooling CNN, we also
consider another model configuration, named as LIN-Pooling, which uses the same model structure as the HOPE-Pooling CNN but removes the orthogonal constraint in eq. (\ref{eq-Upenalty}).
Moreover, the combination of LIN-Input and LIN-Pooling is also used as another baseline for comparison.


In all experiments, we use the mini-batch SGD with a batch size of 100 images to perform 400 epochs of network training. The initial learning rate is $0.06$, and the learning rate should be halved after every 25 epochs. We also use momentum of 0.9 and  weight decay rate of 0.0005. In batch normalization \cite{ioffe2015batch}, we set $\epsilon = 0.001$. For the HOPE-Input and HOPE-Pooling layers, we use a initial $\beta$ that equals to $0.15$, and the $\beta$ should be divided by $1.75$ after every 25 epochs. All weights in CNNs will be initialized by using the method proposed by He et al\cite{he2015delving}. Note that we will not use any data augmentation in this work.

\subsection{Learning Speed}

We firstly consider the computational efficiency of the proposed HOPE methods in learning CNNs. Our computing platform includes Intel Xeon E5-1650 CPU (6 cores), 64 GB memory and a Nvidia Geforce TITAN X GPU (12 GB memory). Our method is implemented with MatConvNet \cite{MatConvNet-2014}, which is a CUDA based CNN toolbox in Matlab. The learning speed of all DCNNs are listed in Table ~\ref{table:time}.


\begin{table}[htb]
  \caption{The learning speed of different DCNNs.}
  \label{table:time}
  \centering
  \begin{tabular}{c|c}
    \toprule
    Methods          &  Learning Speed \\
    \midrule
    Baseline         &   220 images/s    \\
    LIN-Input       &   206 images/s    \\
    HOPE-Input       &   203 images/s    \\
    LIN-Pooling     &   211 images/s    \\
    HOPE-Pooling     &   208 images/s    \\
    LIN-Input + LIN-Pooling & 195 images/s \\
    HOPE-Input + HOPE-Pooling & 190 images/s \\
    \bottomrule
  \end{tabular}
\end{table}

From Table ~\ref{table:time}, we can see that using the more complicated HOPE layers in CNNs only slightly slow down the computation of CNNs in GPUs. Moreover, the learning speed of the HOPE methods is similar with the corresponding LIN methods, which implies that the computational overhead for the orthogonal projection constraint is negligible in training.

\subsection{Performance on CIFAR-10 and CIFAR-100}

We use the classification error rate on the validation sets of the selected databases to evaluate the performance of all CNN models. Besides the 7 CNN configurations we mentioned above, we also include several well-known CNN models from the previous work to compare with our methods, including Tree-Pooling \cite{lee2015generalizing}, BinaryConnect \cite{courbariaux2015binaryconnect} (the performance on CIFAR-100 is not provided), Spectral Pooling \cite{rippel2015spectral}, R-CNN \cite{liang2015recurrent}, Fractional Maxpooling \cite{graham2014fractional} (the performance on CIFAR-10 without data augmentation is not provided), ALL-CNN \cite{springenberg2014striving}, Maxout Networks \cite{goodfellow2013maxout} and Network in Network \cite{lin2013network}.

From all results summarized in Table \ref{table:performance}, we can see that the proposed HOPE-based CNNs models work well in both data sets. And the proposed CNN model that combines HOPE-Input and HOPE-Pooling can achieve the best performances on both CIFAR-10 and CIFAR-100, which are also the state-of-the-art performance
when data augmentation is not used in training. Moreover,
we can see that both HOPE-Input and HOPE-Pooling CNNs consistently outperform the counterpart LIN models that do not use the orthogonal constraints. This implies that the orthogonality introduced by the HOPE methods is quite useful to
improve the performance of CNNs in both image classification tasks. 



\begin{table*}[htb]
  \caption{The classification error rates of all examined CNNs on the validation set of CIFAR-10 and CIFAR-100 (without using data augmentation).}
  \label{table:performance}
  \centering
  \begin{tabular}{c|c|c}
    \toprule
                        & CIFAR-10 & CIFAR-100       \\
    \midrule
    Baseline            & 8.30\%   & 30.71\%         \\
    LIN-Input          & 7.97\%   & 30.13\%         \\
    HOPE-Input          & 7.81\%   & 29.96\%         \\
    LIN-Pooling        & 8.30\%   & 31.85\%         \\
    HOPE-Pooling        & 8.21\%   & 30.60\%         \\
    LIN-Input + LIN-Pooling & 8.22\%        & 31.55\%       \\
    HOPE-Input + HOPE-Pooling & {\bf 7.57\%}  & {\bf 29.80\%}  \\
    \midrule
    Tree-Pooling \cite{lee2015generalizing}  & 7.62\% & 32.37\%  \\
    BinaryConnect \cite{courbariaux2015binaryconnect} & 8.27\% & - \\
    Spectral Pooling \cite{rippel2015spectral} & 8.60\% & 31.60\% \\
    R-CNN \cite{liang2015recurrent}         & 8.69\%  & 31.75\% \\
    F-maxpooling \cite{graham2014fractional} & -       & 31.20\% \\
    ALL-CNN \cite{springenberg2014striving}  & 9.08\%  & 33.71\% \\
    Maxout \cite{goodfellow2013maxout}       & 11.68\% & 34.54\% \\
    Network in Network \cite{lin2013network} & 10.41\% & 35.68\% \\
    \bottomrule
  \end{tabular}
\end{table*}





\section{Conclusions}
\label{Conclusion}

In this paper, we have proposed several methods to apply the recent HOPE model to CNNs for image classification. We have analyzed the relationship between the CNNs and HOPE model, and found a suitable way to use the HOPE method to replace  the convolution and pooling layers in CNNs.
Experimental results on the CIFAR-10 and CIFAR-100 data sets have shown that our proposed HOPE methods work well with CNNs, and can yield the state-of-the-art classification performance in these two data sets.
This study has confirmed that the orthogonal constraints imposed by the HOPE models can significantly improve the performance of CNNs in these image classification tasks.

{\small
\bibliographystyle{unsrt}
\bibliography{HOPECNN}
}

\end{document}